%% file: main.tex
\documentclass[11pt,a4paper]{article}
\usepackage[T1]{fontenc}
\usepackage[utf8]{inputenc}
\usepackage{lmodern}
\usepackage[a4paper,margin=27mm]{geometry}
\usepackage{microtype}
\usepackage{graphicx}
\usepackage{natbib}
\usepackage{tikz}
\usetikzlibrary{arrows.meta,positioning,shapes.geometric,calc}
\usepackage{comment}
\usepackage{float}
\usepackage[table,xcdraw]{xcolor}
\usepackage{colortbl}
\usepackage{booktabs}
\usepackage{caption}
\usepackage{enumitem}
\usepackage{hyperref}
\hypersetup{colorlinks=true,linkcolor=blue!45!black,citecolor=blue!45!black,urlcolor=blue!50!black,pdftitle={CRAFT-AI: Capability and Requirement Alignment for Flexible Manufacturing through Agentic AI},pdfauthor={Kiavash Fathi and Hans Wernher van de Venn}}
\setlist{itemsep=2pt,topsep=4pt}
\title{\textbf{CRAFT-AI:} Capability and Requirement Alignment for Flexible Manufacturing through Agentic AI}
\author{Kiavash Fathi\qquad Hans Wernher van de Venn\thanks{Corresponding author: \href{mailto:wernher@eiras.eu}{wernher@eiras.eu}}\\[7pt]
\small European Institute for Robotics, Automation \& Resilient Systems (EIRAS)\\
\small Winterthur, 8409 Switzerland\\
\small \href{mailto:kiavash@eiras.eu}{kiavash@eiras.eu} \quad | \quad \href{mailto:wernher@eiras.eu}{wernher@eiras.eu}}
\date{\small Preprint}
\begin{document}
\maketitle
\vspace{-1.0em}
\begin{abstract} 
One of the fundamental needs for reaching resilient production infrastructure is flexible manufacturing, which requires translating changing
production requirements into implementable system
configurations. This paper introduces \textbf{CRAFT-AI}, an
agentic AI approach combining natural language (NL)
requirement interpretation, Asset Administration Shell
(AAS)-based capability descriptions, and reusable
engineering knowledge from existing manufacturing
configurations. A tool-using AI agent matches
requirements with available capabilities and
reference implementations to generate manufacturing
scenes and corresponding AAS instances. 
The approach is demonstrated using a Factory I/O prototype with twelve reference scenes and five illustrative manufacturing tasks of varying complexity, providing an initial feasibility assessment of the proposed AAS-enabled, system-level generative engineering approach.
For the four valid user task descriptions, proper requirements, correct references to the relevant engineering knowledge, correct components and lastly a valid Factory I/O scene are generated. For the task with missing engineering knowledge and a wrong sensor selection, the user is informed about which requirements of the scene generation lack domain knowledge and also which sensor is not compatible with the selected raw material by the user with link to the related documentation. After fixing the task description, the correct scene is seamlessly generated. \textbf{CRAFT-AI} highlights the importance of safeguards for risk perception and communication in foundation model-driven solutions.
\end{abstract}
\noindent\textbf{Keywords:} Agentic AI, Flexible Manufacturing, Asset Administration Shell, Requirement and Capability Matching, Resilient Manufacturing Infrastructure, Generative Engineering.
\bigskip
\section{Introduction}\label{sec:introduction}
Manufacturing systems increasingly operate under changing product requirements,
smaller batch sizes, and disruptions that demand rapid reconfiguration of
production resources. Beyond product customization, supply-chain disturbances
can require manufacturers to reconsider both the resources employed and the
configurations in which they operate. The ability to safely and reliably reconfigure production resources in response to changing demands and disruptions is therefore a fundamental prerequisite for resilient manufacturing infrastructure~\citep{puchkova2020resilient, kechagias2026reconfigurable}. In that regard, Manufacturing-as-a-Service (MaaS) has been investigated as one mechanism for improving flexibility and
resilience across production networks \citep{Chen2023FourKnow,Sidorenko2026RAASCEMAN}.
At the factory level, however, responding to new demands still entails the
engineering task of translating often informal objectives into resource
selections and implementable system (re)configurations
\citep{Keddis2015Adaptable,Elbasheer2026NaturalLanguage}.

Capability-based engineering addresses part of this challenge by decoupling
required functions from specific manufacturing resources. In the
product--process--resource perspective, a \emph{required capability}, also referred to as \emph{requirement}, describes
what a production task demands, whereas an \emph{offered capability}, also referred to as \emph{capability}, describes
a function an arbitrary resource can provide. Moreover, the capability--skill--service literature
further distinguishes \emph{capabilities} as implementation-independent functional descriptions
from \emph{skills}, which in the related literature is referred to as their executable implementations
\citep{Eichenwald2024Ontology,Wagner2025Workpieces}.

A practical difficulty remains: the presence of a suitable capability in a
resource catalog does not, on its own, establish that a compatible
implementation, configuration, and sufficient engineering knowledge are
available for a specific production environment. Existing research provides substantial foundations for resolving different
parts of this problem. \citet{Keddis2015Adaptable} map product requirements
and resource capabilities to executable action sequences;
\citet{Kocher2024SMT} generate a formal planning problem from a semantic
capability model.
More recently, \citet{Shi2026Cognitive} combine Asset Administration Shell
(AAS) descriptions and large language models (LLMs) for production
requirement--resource capability matching. At the implementation level,
\citet{Wagner2025Workpieces} address feasibility information for chaining
machining skills using AASs and in-process workpieces. 

Together, these
contributions demonstrate how structured capability descriptions can
support requirement-driven resource selection and
feasibility assessment. Building on these foundations, the present work
investigates how agentic AI can additionally incorporate domain knowledge from
previously implemented manufacturing configurations, referred to as \textit{engineering knowledge}, to bridge the gap between requirements and capabilities matching and translating them into concrete implementable engineering solutions.

LLMs and tool-using agents provide complementary means of interpreting
natural language (NL) tasks and generating engineering artifacts.
\citet{Ni2026LLMAPM} propose LLM-based process planning with state machine
verification; \citet{Korth2026Multiagent} combine multiple agents with
simulation and optimization for production planning.
At the system-model level, \citet{Elbasheer2026NaturalLanguage} generate
simulation models from NL requests, while
\citet{Wang2026Layouts} investigate agent-driven reconfigurable manufacturing
layouts. Thus, neither NL interpretation nor system-level
generation and verification are absent from the literature. 

The pertinent question is instead how to create valid system-level digital twin prototypes with \emph{explicit
engineering knowledge}. Concretely, this refers to converting requirement and capability matches into valid solutions that are in accordance with compatibility, prerequisite, working conditions, \textit{etc.}, of the chosen skills~\citep{grieves2016digital}. 
To address this question, the current paper presents \textbf{CRAFT-AI}, an agentic, requirement-driven
prototype for generating Factory I/O manufacturing scenes using engineering knowledge sources (Fig.~\ref{fig:craft-overview}). A brief demonstration of \textbf{CRAFT-AI} is available \href{https://youtu.be/PjiwWkpq-ik}{online}.
\input{Figures/Fig1} 
In our solution, the NL task description from the user is translated into different requirements by the AI agent. Thereafter, an AAS Type catalog which describes the available assets along with their capabilities, technical properties, \textit{etc.}, is used to address the requirements. Afterwards, a collection of reference Factory I/O scenes, as a set of functioning digital twin prototypes, is used to generate valid Factory I/O scenes. 

During the process of requirement extraction, matching of requirements and capabilities, reference to engineering knowledge and also final solution validation, different artifacts are generated for enhanced traceability and reproducibility.  
Furthermore, a candidate scene without a supported engineering knowledge or not compatible or with wrong capabilities must be
flagged and reported instead of being aborted for creating a starting point for a valid solution in collaboration the user. In our setting, the user can adapt the task description, provide more engineering knowledge, \textit{etc.}, to improve the initially proposed solution by \textbf{CRAFT-AI}. This interaction as a result, transforms our approach into a human-in-the-loop requirement capability matching tool as opposed to a generic agentic AI artifact generator.

The contributions of \textbf{CRAFT-AI} are 3-fold:

\begin{enumerate}
\item {Requirement and capability-driven agentic engineering workflow for converting NL task descriptions into digital twin prototypes.} 
\item {Engineering knowledge-grounded capability selection mechanism for valid solution generation.} 
\item {Traceable generation and checking procedure for improved traceability, reproducibility and facilitating debugging.} 
\end{enumerate}

The remainder of this paper defines the problem, introduces the proposed
architecture and workflow, describes the prototype deployment, presents its
evaluation, and discusses the findings and limitations.

\section{Methodology}\label{sec:methodology}
Building on existing research in generative engineering~\cite{Graessler2023Generative, Koring2026Framework}, we propose a system-level generative engineering framework for resilient manufacturing systems, which we define as follows:

\emph{
\textbf{Generative Engineering} is an iterative, requirement-driven approach to the AI-assisted synthesis, evaluation, and refinement of engineering solutions. It combines structured representations of available resource capabilities, customer requirements, and reusable engineering knowledge to generate candidate system configurations. These configurations are evaluated against functional and quantitative performance criteria using simulation and other verification methods. The resulting feedback guides the iterative refinement of the solutions until the specified acceptance criteria are satisfied or the process is terminated.
}

A key architectural principle of the proposed approach is its application independence. While engineering requirements are inherently application-specific, capabilities are represented through standardized, semantically defined AAS descriptions, independently of their concrete implementation. The core engineering methodology therefore remains application-independent, while domain-specific engineering knowledge, constraints, and capabilities are used to generate application-specific system solutions.

The definition above describes the envisioned end-to-end framework of system-level generative engineering. The present \textbf{CRAFT-AI} prototype implements its initial stages, including NL requirement grounding, AAS-based capability matching, configuration synthesis, and structural assurance. These stages enable the generation of traceable manufacturing configurations and corresponding AAS instances. However, the current implementation does not yet perform automated functional verification through simulation execution or autonomously improve generated solutions based on operational feedback. Simulation-based verification, closed-loop refinement, and systematic learning from engineering experience therefore remain subjects of further research.

In this context, requirements specify the intended functionality
of the manufacturing system, while capabilities describe what an asset, as a technical resource, can do without reducing it to a specific implementation of it. In the current paper, examples of capabilities, include but are not limited to \emph{transfer}, \emph{weigh}, \emph{detect material}, \textit{etc.} Furthermore, skill is an executable deployment of a given capability. This determines how a given functionality can be attained in the production system through its realization. Lastly, engineering knowledge describes how different assets in the existing technical solutions are configured, arranged, parametrized and connected with each another~\citep{verhagen2012critical, Eichenwald2024Ontology,Wagner2025Workpieces, zheng2021application}.

\medskip
\noindent\textbf{Problem formulation:}
Given a NL manufacturing task, a catalog of
available resource capabilities, and a set of reusable
engineering knowledge, the objective is to generate
a system configuration that satisfies the specified
requirements while respecting the available engineering
knowledge and the constraints of the target environment.

The objective is not necessarily to identify a globally
optimal configuration. At the system level, the number
of possible resource combinations, spatial arrangements,
and interface dependencies can result in a large
configuration space. Instead, the initial objective is
to obtain a traceable, technically plausible candidate
that satisfies the specified requirements and can be
evaluated by a human engineer.

If requirements are contradictory, suitable capabilities
are unavailable, or implementation knowledge is
insufficient, the resulting limitations must be explicitly
reported. A partially generated configuration may
provide useful engineering information, but it must not
be presented as a fully validated solution~\citep{freuder1992partial}.

\medskip
\noindent\textbf{Information dependencies:}
Generating a manufacturing configuration requires
information at different levels of abstraction.
The proposed approach distinguishes between the
following complementary information sources:

\begin{enumerate}
\item \textbf{Capability information:}
AAS Type descriptions
provide structured information about resource types
and their available functionality~\citep{grangel2016towards}. Relevant submodels
include:
\begin{itemize}
    \item \texttt{FactoryIOComponent}, which describes
component-specific properties and roles;
    \item \texttt{CapabilityDescription}, which contains declared
capabilities where available; and
    \item \texttt{TechnicalData}, which provides additional
technical information where defined.
\end{itemize}

\item \textbf{Engineering knowledge as implementation information:}
Previously implemented reference scenes contain
concrete component representations, configuration
parameters, spatial arrangements, identifiers, and
interface structures. These examples provide
implementation-specific information that cannot
necessarily be derived from abstract capability
descriptions alone.

\item \textbf{AAS Instance-specific information:}
The generated system must assign unique identifiers,
component names, positions, orientations, and
appropriate configuration values to the selected
resources. These properties describe individual
instances of the selected component types and
establish the relationship between the manufacturing
configuration and its generated AAS instances.
\end{enumerate}

The dependencies between these information sources
vary according to the component functionality.
For example, selecting a conveyor requires a suitable
transport capability, whereas its integration into
the manufacturing system additionally requires
information about its physical placement and
configuration. Similarly, selecting a sensor based
on its detection capability is insufficient unless
its position and relevant input/output identifiers
are correctly represented. These dependencies
must be considered when generating a coherent
system configuration.

\medskip
\noindent\textbf{Representation constraints:}
In the presented prototype, Factory I/O serves as
the target digital twin prototype environment. Its scene files
are represented using XML and contain both
scene-level information and individual object
descriptions. An object is associated with a
\texttt{PrefabName}, which identifies a predefined
Factory I/O component type. The corresponding
object representation also contains the information
needed for its instantiation and configuration.

Conceptually, the relevant information in a scene
can be represented as follows:

\footnotesize
\begin{verbatim}
Scene:
  - Version and scene metadata
  - Object 1
      - Instance identifier
      - PrefabName
      - Position and rotation
      - Component configuration
      - I/O, tags and references
  - Object 2
      - ...
\end{verbatim}

\normalsize
This representation illustrates the logical
information dependencies rather than the exact
XML schema. The actual serialization is obtained
from the existing reference scenes.

A prefab provides a predefined component
representation within the Factory I/O environment.
However, the corresponding \texttt{PrefabName}
alone is not sufficient to generate a valid scene
object. The complete object structure, its internal
configuration, and the consistency of referenced
identifiers must also be preserved.

The central challenge is therefore to combine
abstract capability information with concrete
engineering knowledge without introducing
unsupported component definitions or invalid
references. Furthermore, the generated AAS
instances must remain consistent with the
components actually included in the resulting
manufacturing configuration. Accordingly, the proposed approach investigates
how a tool-using AI agent can bridge these
information layers to generate traceable
manufacturing configurations from NL
requirements. Structural validation provides an
initial consistency check, while functional
correctness and operational behavior require
separate verification in the target environment.
\section{Solution Overview}
\label{sec:solution-overview}
Our \textbf{CRAFT-AI} approach combines NL
requirement interpretation, AAS-based capability matching,
and reusable engineering knowledge within an agentic
engineering workflow. As illustrated in
Fig.~\ref{fig:prompt-structure}, the workflow is organized
into four phases: grounding, matching, synthesis, and
assurance. A structured prompt defines ten processing
steps, while a tool-using AI agent executes the engineering
tasks using the available information sources.
\input{Figures/Fig2}

\paragraph{Structuring capabilities and skills as an AAS catalog:}
For having a single source of truth for different capabilities and the related skills, AAS representation of the relevant information from different sources of unstructured data is created. This so-called AAS Type catalog provides structured component descriptions, including technical properties
and offered capabilities. The AAS catalog is developed
using ChatGPT based on the unstructured data in \href{https://docs.factoryio.com/manual/parts/}{Factory I/O documentation} and
the standardization guidelines from the \href{https://industrialdigitaltwin.org/en/content-hub/submodels}{IDTA submodel templates}, followed by refinement performed by an AAS domain expert. 

\paragraph{Engineering knowledge for digital twin prototyping:}
\href{https://docs.factoryio.com/manual/scenes/}{The twelve existing
manufacturing simulation scenes from Factory I/O} represent reusable
engineering knowledge, providing component configurations, object structures, and interface
information. While the AAS catalog describes what
resources can potentially accomplish, the reference
scenes provide implementation knowledge describing
how components should be configured.

\paragraph{Requirement \& capability matching:}
The agent decomposes the user-provided NL
task into individual requirements with unique identifiers.
These requirements are matched against the capabilities and skills
offered by the AAS Type catalog to identify suitable
resources. Each selection is documented with its
corresponding requirement, AAS Type, matched
capabilities, candidate skills, and also justification. Subsequently, the
agent verifies whether a corresponding implementation from the chosen skills
exists in the reference scenes. Components without
a valid reference object are excluded to prevent
unsupported configurations. This combines functional
capability matching with experience-based
implementation constraints.

\paragraph{Digital twin prototype generation and validation:}
Based on the selected components, the agent generates
a scene plan defining component instances, positions,
orientations, and configurations. Reference objects
are reused to construct the manufacturing scene,
while corresponding AAS instances maintain the
relationship between component types and their
generated representations. Intermediate artifacts,
including requirements, component selections,
and the scene plan, preserve decision traceability.
Static validation checks structural consistency,
identifiers, reference provenance, and requirement
coverage. However, functional correctness in Factory I/O environment requires
separate runtime verification.

\paragraph{Human--AI interaction:}
 A \href{https://streamlit.io/}{Streamlit}-based interface enables users to explore the AAS catalog, inspect available capabilities and skills, specify preferred components, and submit manufacturing tasks in NL. The application invokes the Codex agent through a Python execution wrapper and subsequently presents the generated engineering artifacts, validation results, and execution logs.

The current implementation follows a human-supervised engineering workflow. While the AI agent performs requirement interpretation, capability matching, and configuration synthesis, the human engineer remains responsible for evaluating the generated results. If required implementation knowledge is unavailable or inconsistencies between selected components are identified, the agent reports the corresponding limitations and provides relevant supporting information. The user can subsequently revise the task description, modify component selections, or provide additional engineering knowledge before initiating another generation cycle.

This interaction establishes an initial human-in-the-loop mechanism for identifying and resolving limitations in generated engineering configurations. However, the current prototype does not autonomously learn from human corrections or systematically incorporate these interactions into an evolving engineering knowledge base. Within the broader system-level generative engineering framework, engineering decisions, expert assessments, corrective interventions, and validated operational feedback are envisioned as sources of reusable engineering knowledge that can support the continuous improvement of subsequent engineering solutions.

\section{Solution Deployment}\label{sec:solution-deployment}
The software architecture and
information flow for \textbf{CRAFT-AI} are illustrated in
Fig.~\ref{fig:craft-software-architecture}.
\input{Figures/Fig3}
The application consists of three main modules.
The \texttt{app.py} module provides the user interface
for exploring available capabilities and skills, submitting
manufacturing requirements, and inspecting generated
results. 

\texttt{factoryio\_catalog.py} 
parses and prepares the AAS Type catalog for user
interaction. The \texttt{run\_codex.py} module combines
the task description with the structured engineering
prompt and invokes the Codex agent within the
project workspace.

During execution, the agent accesses the AAS catalog
and engineering knowledge to generate manufacturing
configurations and corresponding AAS instances for the skills.
Intermediate artifacts, validation reports, and
execution logs are stored in dedicated directories
to support traceability and debugging. An additional
utility, \texttt{archive\_test.py}, enables the
archiving of individual test runs for subsequent
analysis and reproducibility.

The prototype operates locally and requires
human inspection of the generated configurations.
Operational validation in Factory I/O remains
separate from the automated static checks.

\section{Results}\label{sec:results}
The \textbf{CRAFT-AI} prototype was examined using five
illustrative manufacturing tasks of varying complexity.
These examples investigate the generation of manufacturing
configurations and the handling of missing engineering knowledge. Given the higher system-level view of this paper, automatic PLC code generation is out of the scope of this work. Therefore, the performance of the proposed solution \textbf{CRAFT-AI} is evaluated by inspecting: (KPI-I Task) if the requirements completely cover the NL task description, (KPI-II \# Reqs) how many of the requirements could be addressed, (KPI-III \# AAS) if all the selected skills have a dedicated AAS instance and (KPI-IV DTP) if a valid digital twin prototype is generated from the NL task description. The aforementioned five illustrative manufacturing tasks are as follows. The results are presented in Table~\ref{tab:craft-ai-kpis}.

\textbf{Scene I \& II - Basic manufacturing scenes:}
The first two examples involve relatively simple
configurations. In the first case, the agent was instructed
to generate \href{https://docs.factoryio.com/manual/parts/items/#product-base}{green bases}, transport them along a \href{https://docs.factoryio.com/manual/parts/light-load/#belt-conveyors}{conveyor
belt}, and remove them at the end of the line. The second case involved
a \href{https://docs.factoryio.com/manual/parts/stations/#tank}{water tank} with dedicated ``Fill'' and ``Empty'' buttons.

\textbf{Scene III - Unit sorting and storage:}
In this scene, mixed \href{https://docs.factoryio.com/manual/parts/items/#boxes}{production units} are sorted based on their \href{https://docs.factoryio.com/manual/parts/sensors/#light-array-emitter-and-receiver}{height}, sent to their corresponding conveyor belt and then \href{https://docs.factoryio.com/manual/parts/stations/#pick-place}{palletized} for efficient storage.

\begin{figure}[t]
    \centering
    \includegraphics[width=0.9\columnwidth]{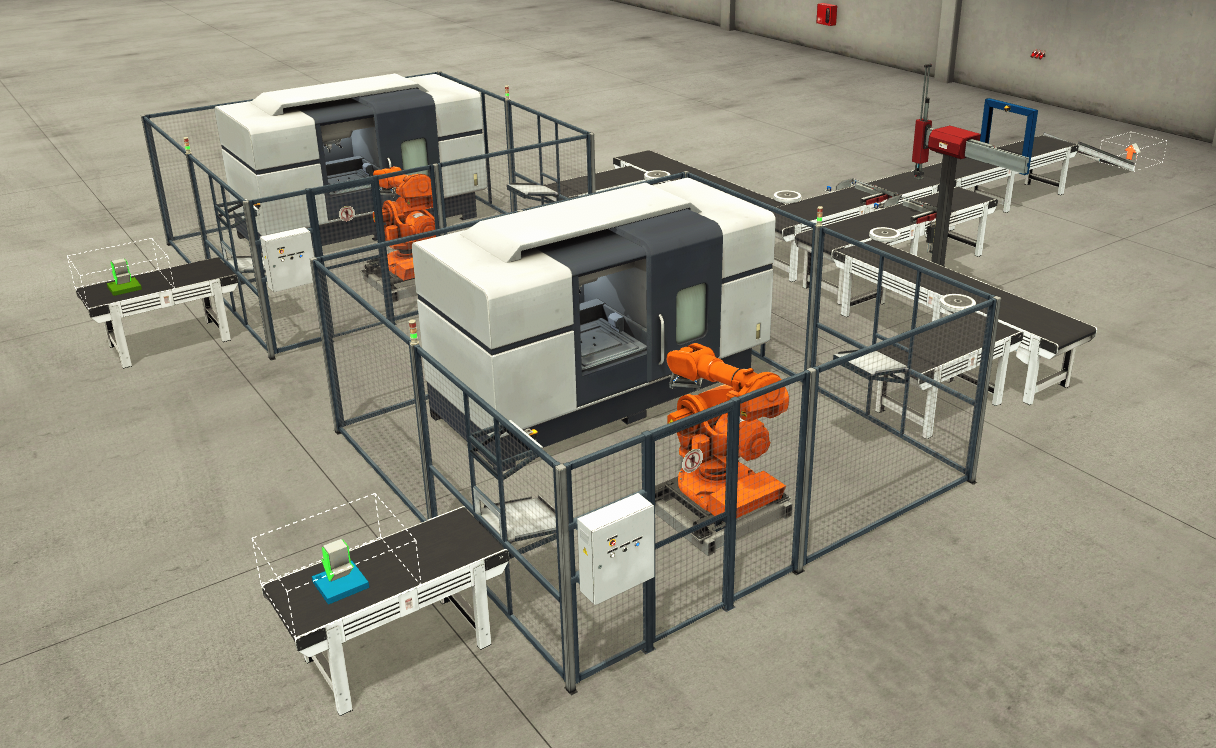}
    \caption{Scene IV generated by \textbf{CRAFT-AI} as an assembly line example.
    }
    \label{fig:CRAFT-AI-example}
\end{figure}

\textbf{Scene IV - Multi-stage assembly line:}
As a more complex task, this case required two production lines to
manufacture green bases and blue lids, transport them
to a \href{https://docs.factoryio.com/manual/parts/stations/#two-axis-pick-place}{pick-and-place} assembly station, \href{https://docs.factoryio.com/manual/parts/sensors/#vision-sensor}{inspect} the
assembled product, and subsequently remove it.
The resulting scene is illustrated in
Fig.~\ref{fig:CRAFT-AI-example}.

\textbf{Scene V - Unavailable engineering skill with wrong instructions:}
The last example requested a raw material-handling configuration involving a curved conveyor selected through the user interface, a pick-and-place station,
and an RFID sensor. Since suitable implementation knowledge for the curved conveyor was unavailable in the reference scenes, the agent
could not generate the complete configuration. 
Additionally, the agent identified that the selected RFID reader was unsuitable for detecting the raw material, since raw material objects in Factory I/O do not contain RFID tags.
The user is also provided a \href{https://docs.factoryio.com/manual/parts/sensors/#rfid-tags}{link to the documentation} to verify this warning issued by the agent and chose a fitting sensor.
After changing the conveyor belt and the sensor for counting the throughput in the NL task description by the user, the scene is successfully generated. 

Overall, the examples illustrate how \textbf{CRAFT-AI} can
combine capability descriptions with reusable
engineering knowledge to generate traceable
manufacturing configurations. They also demonstrate
a limitation of the approach: successful generation
depends on sufficient reference knowledge, indicating that
structural consistency alone does not establish
functional correctness.

\begin{table}[t]
\centering
\small
\setlength{\tabcolsep}{6pt}
\renewcommand{\arraystretch}{1.15}

\caption{CRAFT-AI performance.}

\begin{tabular}{@{}lcccc@{}}
\toprule
\textbf{Scene} &
\shortstack{\textbf{KPI-I}\\Task} &
\shortstack{\textbf{KPI-II}\\\# Reqs} &
\shortstack{\textbf{KPI-III}\\\# AAS} &
\shortstack{\textbf{KPI-IV}\\DTP} \\
\midrule
I       & Yes & 3/3 & 4/4 & Yes \\
II      & Yes & 4/4 & 3/3 & Yes \\
III     & Yes & 6/6 & 32/32 & Yes \\
IV      & Yes & 9/9 & 26/26 & Yes \\
\addlinespace[3pt]
V ($1^{st}$ Try) & Yes & 5/7 & 5/7 & No \\
V ($2^{nd}$ Try) & Yes & 7/7 & 9/9 & Yes \\
\bottomrule
\end{tabular}
\label{tab:craft-ai-kpis}
\end{table}

\section{Conclusion} \label{sec:conclusion}

This paper presented \textbf{CRAFT-AI}, an agentic AI approach combining NL requirement interpretation, AAS-based capability descriptions, and reusable engineering knowledge to generate traceable manufacturing configurations and corresponding digital twin prototypes.

The proposed approach builds on a fundamental separation between application-specific requirements, semantically described resource capabilities, and reusable engineering knowledge. This separation establishes an application-independent architectural foundation for system-level generative engineering, in which a common engineering methodology can be applied to generate solutions for different industrial applications.

The feasibility of the approach was investigated through five illustrative manufacturing tasks using Factory I/O as the target environment. The results demonstrated that \textbf{CRAFT-AI} can translate NL tasks into structured requirements, identify suitable capabilities, reuse implementation knowledge, and generate structurally consistent manufacturing configurations with corresponding AAS instances. All five tasks ultimately achieved complete requirement coverage and successful prototype generation. The initial unsuccessful attempt in the fifth scenario further highlighted the importance of identifying missing engineering knowledge and incompatible component selections, enabling the human user to revise the task and initiate a subsequent generation cycle.

The current implementation addresses the initial stages of system-level generative engineering, including requirement grounding, capability matching, configuration synthesis, and structural assurance. However, structural consistency alone does not establish functional correctness or operational feasibility. Automated simulation-based verification, quantitative performance evaluation, and closed-loop improvement remain important subjects of further research.

Future work will therefore focus on integrating functional verification and iterative optimization into the generative engineering workflow, investigating the systematic reuse of knowledge obtained from engineering decisions and operational feedback, and evaluating the practical applicability of the proposed architecture across different industrial domains.

Ultimately, the research aims to advance system-level generative engineering towards a knowledge-driven, capability-based, and AI-assisted methodology for the development of intelligent industrial systems.

\section*{Acknowledgements}
The authors gratefully acknowledge \href{https://realgames.co/}{Real Games} for their support of this research.
\vspace{0.2cm}

\noindent 
This manuscript is a preprint intended for submission to the \textbf{37th European Safety and Reliability Conference (ESREL 2027)}, taking place on 6–10 June 2027 at ETH Zürich, Switzerland, for peer review and publication in the conference proceedings.

\section*{Generative AI Disclosure}
During the preparation of this work the authors used ChatGPT in order to summarize information, generate figures and process text. The authors reviewed and edited the content and take full responsibility for it.

\bibliographystyle{chicago}
\bibliography{References}

\end{document}

%% file: Figures/Fig1.tex
\begin{figure}[t!]
    \centering
    \includegraphics[width=0.75\linewidth]{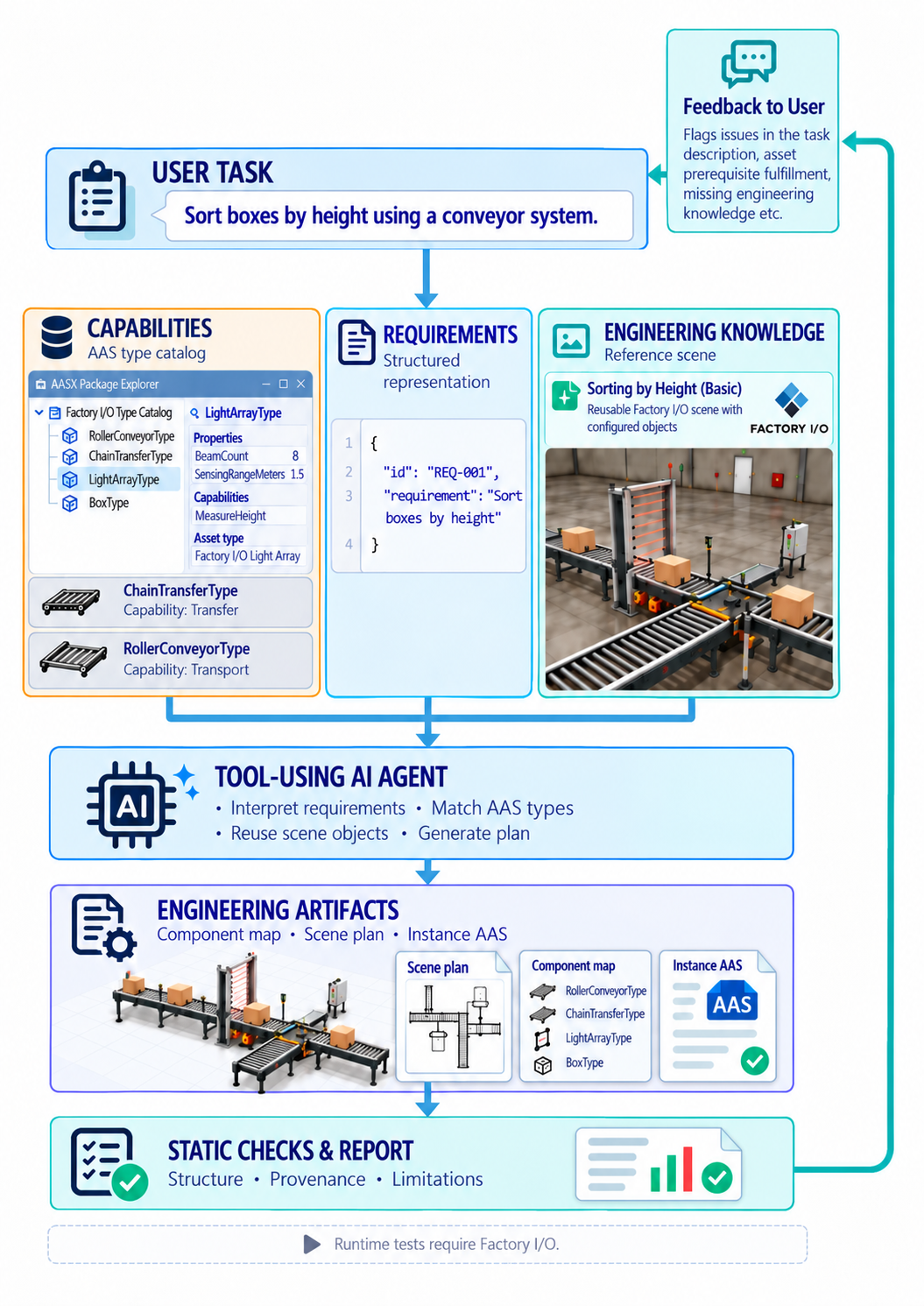}
    \vspace{-0.9cm}
    \caption{Overview on \textbf{CRAFT-AI}.
    \vspace{-0.6cm}
    }
\label{fig:craft-overview}
\end{figure}

%% file: Figures/Fig2.tex
\begin{figure*}
    \centering
    \includegraphics[width=1\linewidth]{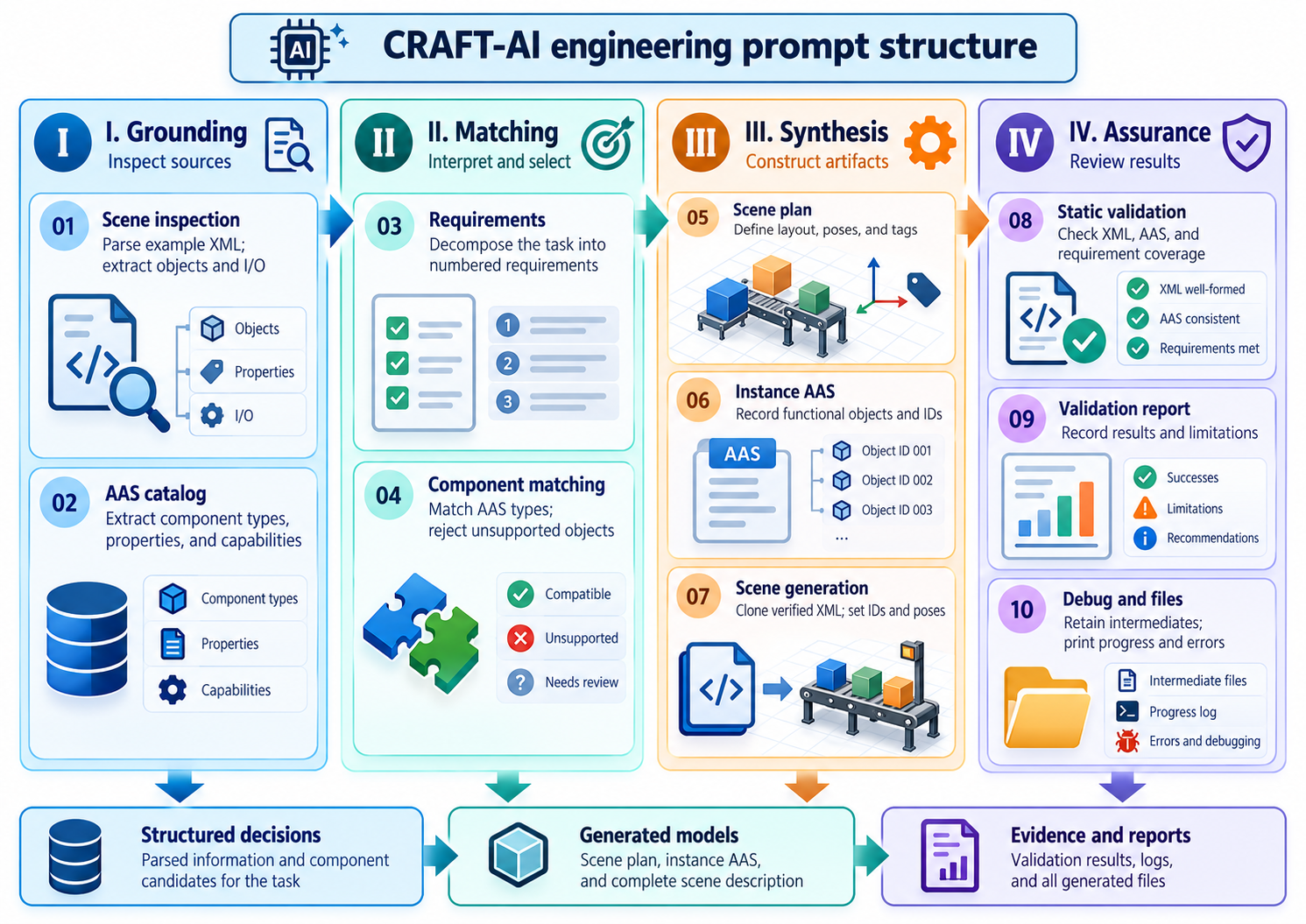}
    \caption{Structure of the engineering prompt provided to the tool-using AI agent.
\vspace{-0.4cm}
}
    \label{fig:prompt-structure}
\end{figure*}

%% file: Figures/Fig3.tex
\begin{figure*}
    \centering
    \includegraphics[width=1\linewidth]{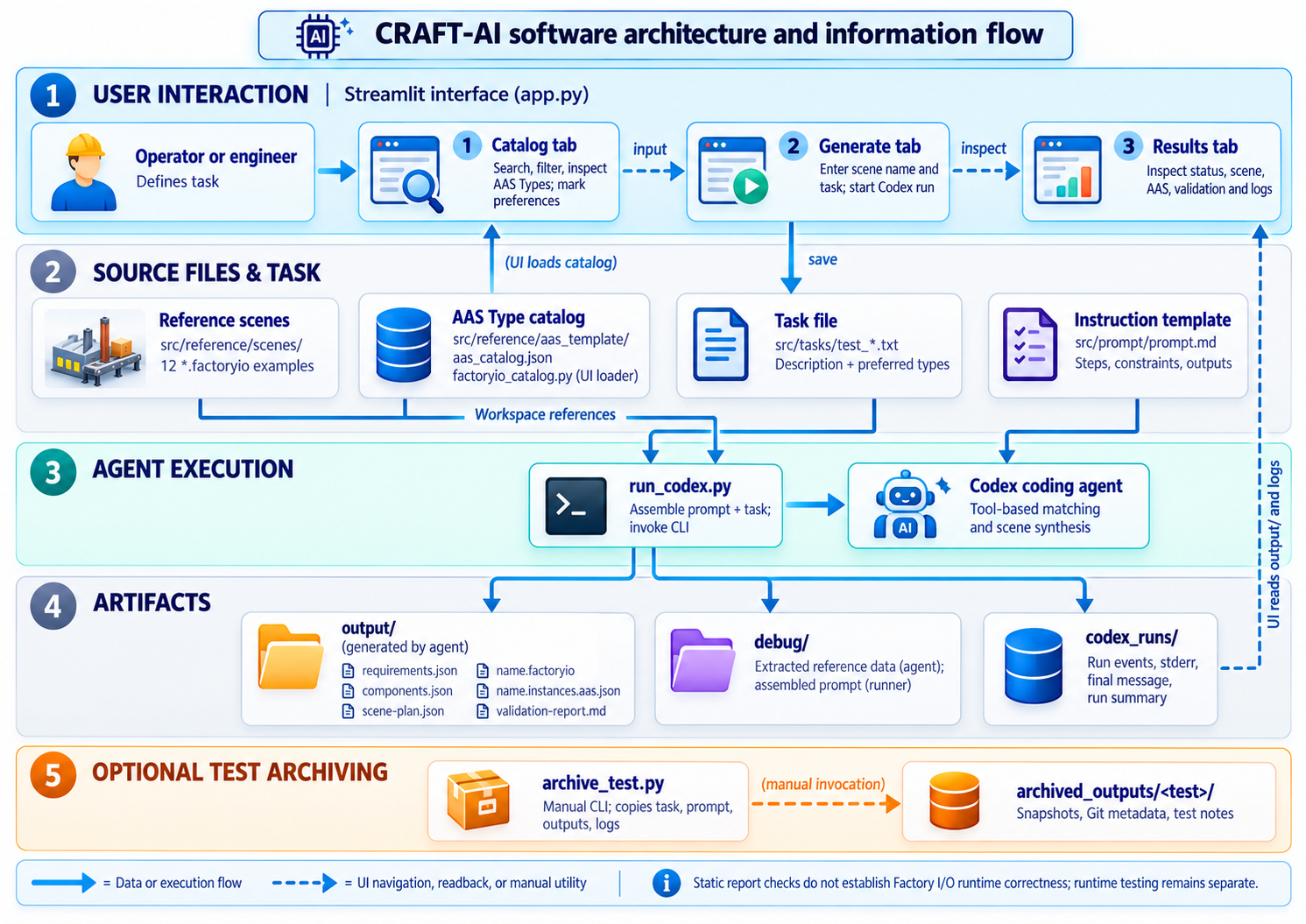}
    \caption{Software architecture and information flow of the CRAFT-AI prototype.
\vspace{-0.4cm}
}
    \label{fig:craft-software-architecture}
\end{figure*}